\documentclass[submission,copyright,creativecommons]{eptcs}
\providecommand{\event}{TARK 2023} 

\usepackage{iftex}

\usepackage{amsmath}
\usepackage{amssymb}
\usepackage{amsfonts}
\usepackage{bbm}
\usepackage{amsthm}
\usepackage{thmtools, thm-restate}

\usepackage{graphicx}

\newtheorem{definition}{Definition}

\ifpdf
  \usepackage{underscore}         
  \usepackage[T1]{fontenc}        
\else
  \usepackage{breakurl}           
\fi

\title{Aggregating Credences into Beliefs:\\ Agenda Conditions for Impossibility Results\\ (extended abstract)}
\author{Minkyung Wang
\institute{CONCEPT\\ Cologne, Germany}
\institute{Department of Philosophy\\
University of Cologne\\
Cologne, Germany}
\email{minkyungwang@gmail.com}
\and
Chisu Kim
\institute{Independent Researcher\\
Cologne, Germany}
\email{\quad tschuessu@gmail.com}
}
\def\titlerunning{Aggregating Credences into Beliefs}
\def\authorrunning{Minkyung Wang \& Chisu Kim}
\begin{document}
\maketitle

\begin{abstract}
Binarizing belief aggregation addresses how to rationally aggregate individual probabilistic beliefs into collective binary beliefs. Similar to the development of judgment aggregation theory, formulating axiomatic requirements, proving impossibility theorems, and identifying exact agenda conditions of impossibility theorems are natural and important research topics in binarizing belief aggregation. Building on our previous research on impossibility theorems, we use an agenda-theoretic approach to generalize the results and to determine the necessary and sufficient level of logical interconnection between the issues in an agenda for the impossibility theorems to arise.
We demonstrate that (1) path-connectedness and even-negatability constitute the exact agenda condition for the oligarchy result stating that binarizing belief aggregation satisfying proposition-wise independence and deductive closure of collective beliefs yields the oligarchies under minor conditions; (2) negation-connectedness is the condition for the triviality result obtained by adding anonymity to the oligarchy result; and (3) blockedness is the condition for the impossibility result, which follows by adding completeness and consistency of collective beliefs. Moreover, we compare these novel findings with existing agenda-theoretic characterization theorems in judgment aggregation and belief binarization.
\end{abstract}

\section{Introduction}

The question of how to rationally aggregate individual beliefs into collective beliefs is important and ubiquitous in our society. In this regard, there has been abundant literature on collective decision theory, judgment aggregation, and probabilistic opinion pooling studies. One of the essential features of belief is that there are different types of beliefs. For example, some beliefs may be represented by traditional "logical" languages---she believes that it is raining outside---while other types of beliefs might be modeled by "probability functions"---she believes with 90 percent certainty that it is raining outside. Logical languages are similar to our natural languages and are therefore efficient for communicating with human agents, despite the fact that they sometimes suffer from significant information reduction, as in the case of the Lottery paradox. In contrast, probabilistic beliefs hold a fair amount of information to deal with uncertain environments, although people usually do not reach that level of precision.
Considering these pros and cons of different types of beliefs, it is not surprising that different types of beliefs may be required at different stages of belief aggregation procedures depending on situations. If objective chances of issues in question can be given, it is epistemically preferable to report individual opinions in terms of degrees of belief. If the conclusion of an epistemic collective decision guides action (e.g., a jury verdict), it is practically better to report the collective opinion by means of plain logic. Therefore, rational belief aggregation should be able to deal with different types of beliefs. 
One important topic in aggregating one type of belief into a different type of belief is aggregating probabilistic beliefs into collective binary beliefs (e.g., \cite{IS19} \cite{WAN22}). We call this subject matter "binarizing belief aggregation" \cite{WAN22}. We can observe these belief aggregation problems in expert panels, the scientific community, and political parties, whenever individuals' opinions can be encoded probabilistically, and the group's beliefs should be more decisive.

Similar to the development of judgment aggregation theory (e.g., \cite{DH10} \cite{NP10}), formulating axiomatic requirements, proving impossibility theorems, and identifying exact agenda conditions of impossibility theorems are natural and important research topics in binarizing belief aggregation. Building on our previous research on impossibility theorems, this paper uses an agenda-theoretic approach to determine which level of logical interconnection between the issues in an agenda is necessary and sufficient for the impossibility theorems to arise.
Indeed, our previous paper assumed the agenda to be an algebra, which is the most typical when dealing with probabilistic beliefs. However, in practice, the agenda being an algebra might be quite demanding because we might not be interested in, for example, the conjunction of two propositions when making a collective decision on the two propositions. Besides the literature on judgment aggregation, agenda-theoretic approaches can be found in other fields as well. In probabilistic opinion pooling, general agendas were investigated to characterize linear pooling (e.g., \cite{DL17a} \cite{DL17b}). In the belief binarization problem, general agendas were studied to characterize impossibility theorems (e.g., \cite{DL18} \cite{DL21}).

In this study, we demonstrate that (1) path-connectedness and even-negatability constitute the exact agenda condition for the oligarchy result, which states that binarizing belief aggregation satisfying proposition-wise independence and deductive closure of collective beliefs yields the oligarchies under certain conditions; (2) negation-connectedness is the condition for the triviality result obtained by adding anonymity to the oligarchy result; and (3) blockedness is the condition for the impossibility result, which follows by adding completeness and consistency of collective beliefs. Moreover, we compare these novel findings with existing agenda-theoretic characterization theorems in judgment aggregation and belief binarization. All proofs of lemmas and theorems are provided in the full paper.

\section{Binarizing Belief Aggregation and the impossibility results} \label{sec2}

We begin by introducing notations and definitions we will use throughout this paper. 
Let $W$ be a finite non-empty set of possible worlds. 
An \textit{agenda} $\mathcal{A}$ is a non-empty set of subsets of $W$ that is closed under complement.
Let $N:=\{1,...,n\}$($n \geq 2$) be the set of individuals.
For each $i \in N$, an individual $i$'s \textit{probabilistic belief} $P_i$ is a function extendable to a probability function on the smallest algebra that includes $\mathcal{A}$.
We denote by $\vec{P}:=(P_1,...,P_n)=(P_i)_{i \in N}$ a profile of $n$ individuals' probabilistic beliefs.
Binarizing belief aggregation deals with individuals' probabilistic beliefs and the group's binary beliefs.
Binary beliefs are represented by a function $Bel:\mathcal{A} \rightarrow \{0 ,1 \}$. 
Sometimes, we abuse the notation and denote by $Bel$ 
the \textit{belief set} $ \{ A \in \mathcal{A} \vert \mbox{ } Bel(A)=1 \} $, and $Bel A$ is a shorthand for $A \in Bel$ or $Bel(A)=1$.
A binarizing aggregator (BA) $F$ is a function that takes a profile $\vec{P}$ of $n$ probabilistic beliefs in a given domain and returns a binary belief $F(\vec{P})$.

Now, let us define the axiomatic requirements on BA that are needed to formulate our impossibility results. 
First, we need the following rationality requirements on the domain and codomain of a BA.

$\bullet$ Universal Domain (UD): the domain of $F$ is the set of all profiles $\vec{P}$ of $n$ probabilistic beliefs

$\bullet$ Collective Deductive Closure (CDC)/Consistency (CCS)/Completeness (CCP): for all $\vec{P}$ in the domain, the resulting collective beliefs $F(\vec{P})$ is deductively closed/consistent/complete, respectively

Note that a binary belief $Bel$ is deductively closed iff it holds that, if $Bel \vDash A (i.e., \bigcap Bel \subseteq A)$, then $Bel A$ for all $A \in \mathcal{A}$. Moreover, $Bel$ is consistent if $Bel \nvDash \emptyset$, and $Bel$ is complete if $Bel A$ or $Bel \overline{A}$ for all $A \in \mathcal{A}$ where $\overline{A}$ is the complement of $A$. Second, we enlist different rationality requirements on BAs themselves.

$\bullet$ Certainty Preservation (CP)/Zero Preservation (ZP):  for all $A \in \mathcal{A}$, if $ \vec{P}(A) (:=(P_1(A), \cdots, P_n(A))) = (1,...,1)$/$ \vec{P}(A) = (0,...,0)$, then $F(\vec{P})(A)=1$/$F(\vec{P})(A)=0$, respectively, for all $\vec{P}$ in the domain of $F$.

$\bullet$ Anonymity (AN): $F((P_{\pi(i)})_{i \in N})$ $=F((P_i)_{i \in N})$ for all $\vec{P}$ in the domain of $F$  and all permutation $\pi$ on $N$.

$\bullet$ Independence (IND): for all $A \in \mathcal{A}$, there exists a function $G_A$ such that $F(\vec{P})(A) = G_{A}(\vec{P}(A))$ for all $\vec{P}$ in the domain of $F$.

$\bullet$ Systematicity (SYS): there exists a function $G$ such that $F(\vec{P})(A) = G(\vec{P}(A))$ for all $A \in \mathcal{A}$ and for all $\vec{P}$ in the domain of $F$. 

Our previous paper \cite{WAN22} proved the following theorems under the assumption that $\mathcal{A}$ is an algebra with at least three elements besides the empty set and W, which we call a non-trivial algebra. We aim to relax this in this study.

1. (The Oligarchy Result) The only BAs satisfying UD, CP, ZP, IND, and CDC are the following oligarchies: there is a non-empty subset $M$ of $N$ such that
\[
F(\vec{P})(A)= \left\{
\begin{array}{rl}
	1    &    \mbox{if }  P_i(A)=1  \mbox{ for all } i \in M  \\
	0    &    \mbox{otherwise} 

\end{array}\right.
\]
for all $A \in \mathcal{A}$.

2. (The Triviality Result) The only BAs satisfying UD, CP, ZP, IND, CDC and AN are the oligarchy with $M=N$, which we call the trivial rule.

3. (The Impossibility Result) There is no BA satisfying UD, CP, IND, CCP, and CCS.

\section{The Agenda Condition for the Oligarchy Result}
\label{sec3}

This section presents and proves our first main result: the agenda condition for the oligarchy result.
The following two agenda conditions have been extensively studied, as they characterize the most famous impossibility agendas in judgment aggregation.

\begin{definition}[Path-connected and Even-negatable Agenda]
(1) For any $A, B \in \mathcal{A}$, we say that 
$A$ conditionally entails $B$ ($A \vDash^{*} B$) if there is a subset $\mathcal{Y} \subseteq \mathcal{A}$ that is consistent with $A$ and $\overline{B}$\footnote{That is, 
$  \mathcal{Y} \cup \{ A \} \nvDash \emptyset$ and 
$  \mathcal{Y} \cup \{ \overline{B} \} \nvDash \emptyset$} such that 
$\{A \} \cup \mathcal{Y} \vDash B$ (i.e., $\bigcap (\{A\} \cup \mathcal{Y}) \subseteq B$ and we write this as $A \vDash_{\mathcal{Y}}^{*} B$).
An agenda $\mathcal{A}$ is path-connected (PC) if $A \vDash^{**} B$ for all contingent issues $A, B \in \mathcal{A}$, where $\vDash^{**}$ is the transitive closure of $\vDash^{*}$.\\
(2) An agenda $\mathcal{A}$ is even-negatable (EN) iff there is a minimally inconsistent set $\mathcal{Y} \subseteq \mathcal{A}$ such that $\mathcal{Y}_{\neg \mathcal{Z}}:= (\mathcal{Y} \setminus \mathcal{Z}) \cup \{ \overline{A} \vert \mbox{ } A \in \mathcal{Z} \}$ is consistent for some subset $\mathcal{Z} \subseteq \mathcal{Y}$ of even size.
\end{definition}

Path-connectedness means that every two issues are connected by a path, i.e., a chain of conditional entailment relations. Regarding conditional entailment relation, let us mention a useful fact. If $A \vDash^{*}_{\mathcal{Y}} B$, it also holds that $\overline{B} \vDash^{*}_{\mathcal{Y}} \overline{A}$, and thus if $A \vDash^{**} B$, then $\overline{B} \vDash^{**} \overline{A}$. 
And even-negatability says that a minimally inconsistent subset of the agenda can be made consistent by negating some even number of its element. It is well-known that an agenda is even-negatable unless the propositions in the agenda are composed only with negation and biconditional from some logically independent propositions.
%
Note that these two conditions are weaker than the agenda being a non-trivial algebra, which is the assumption on the agenda in \cite{WAN22}. 

\begin{restatable}[]{lemma}{nontrivial}
\label{nontrivial}
Every non-trivial algebra is path-connected and even-negatable.
\end{restatable}
From now on, we add one more assumption on $\mathcal{A}$ that $\emptyset \notin \mathcal{A}$(and thereby $W \notin \mathcal{A}$).\footnote{In the following, especially in Theorem \ref{thm:Agenda Conditions for Triviality Result} and Theorem \ref{thm:Agenda Conditions for Impossibility Result}, we will use some results of Nehring \& Puppe (2010), where the agenda consists of contingent issues.
To describe our proof more simply, we adopt that assumption.
} 
Thus, our agenda $\mathcal{A}$ is a complement-closed finite non-empty set of some contingent subsets of the underlying set $W$. 
The following lemma shows that path-connectedness is sufficient to obtain what is called the contagion lemma.

\begin{restatable}[Agenda Condition for the Contagion Lemma]{lemma}{pathandsys}
\label{lem:path-connected IND to SYS}
Let $\mathcal{A}$ be path-connected. If a BA $F$ with UD satisfies CDC, CP, and IND, then it satisfies SYS.
\end{restatable}


This lemma parallels the one in generalized opinion pooling of Dietrich \& List (2017a): path-connectedness characterizes that if generalized OP satisfies CP and IND, then it satisfies SYS.
In our lemma as well, its converse---if $\mathcal{A}$ is not path-connected, then there is a BA F on $\mathcal{A}$ satisfying CDC, CP, and IND but not SYS---also holds.
The counterexample will be indicated in the proof of Theorem \ref{thm:Agenda Oligarchy}.

The following definition and lemma will be needed to prove our succeeding main theorem. 

\begin{definition}[Non-simple Agenda and Pair-negatable Agenda] 
(1) An agenda $\mathcal{A}$ is non-simple(NS) iff there is a minimally inconsistent subset $\mathcal{Y} \subseteq \mathcal{A}$ with $\vert \mathcal{Y} \vert \geq 3$.

(2) An agenda $\mathcal{A}$ is pair-negatable iff there is a minimally inconsistent set $\mathcal{Y} \subseteq \mathcal{A}$ such that $\mathcal{Y}_{\neg Z}$ is consistent for some subset $\mathcal{Z} \subseteq \mathcal{Y}$ with $\vert \mathcal{Z} \vert = 2$.
\end{definition}

Non-simple agendas can be used as a criterion for determining whether a given agenda has minimal complexity. Pair-negatable agendas are a special case of even-negatable agendas.
The following lemma shows that a pair-negatable agenda is sufficient to be an even-negatable agenda, and a path-connected agenda already has a fairly complex structure. 

\begin{restatable}[]{lemma}{pairnegatable}
\label{lem:pair-neg and non-simple}
(1) An agenda $\mathcal{A}$ is even-negatable iff $\mathcal{A}$ is pair-negatable.

(2) If an agenda $\mathcal{A}$ is path-connected, then it is non-simple.
\end{restatable}


Now we prove that the agenda being path-connected and even-negatable is the sufficient and necessary condition for the oligarchy result.

\begin{restatable}[Agenda Condition for the  Oligarchy Result]{theorem}{agendaoli}
\label{thm:Agenda Oligarchy}
Let $\vert N \vert \geq 3$.
An agenda $\mathcal{A}$ is path-connected and even-negatable iff the only BAs on $\mathcal{A}$ satisfying UD, ZP, CP, IND, and CDC are the oligarchies.

\end{restatable}

The only-if direction of the theorem generalizes the oligarchy result and shows that even if an agenda satisfies a weaker condition---path-connectedness and even-negatability---than a non-trivial algebra, the oligarchy result holds. 
If we examine the proof of the oligarchy result in \cite{WAN22} in detail, we can observe that the agenda condition was used solely to establish the following two facts:\\ 
(Fact 1)  if $\vec{a} \leq \vec{b}$ and if $G(\vec{a})=1$, then $ G(\vec{b})=1$ where $G$ is a function satisfying $F(\vec{P})(A)=G(\vec{P}(A))$.\\ 
(Fact 2) if $\vec{a}+\vec{b}-\vec{1} \geq \vec{0}$ and if $G(\vec{a})=1$ and $ G(\vec{b})=1$, then $G(\vec{a}+\vec{b}-\vec{1} )=1$.\\ 
Therefore, to prove the only-if direction, it is enough to derive (Fact 1) from even-negatability and (Fact 2) from path-connectedness. The agenda conditions are only relevant to (Fact 1) and (Fact 2), and once we see that they hold then we can apply the proof of the oligarchy result in \cite{WAN22}.

Our proof also reveals that if we assume the stronger property of SYS instead of IND, then Lemma \ref{nontrivial} is not needed, and non-simplicity (NS) is sufficient to obtain the oligarchy result.
This observation indicates that stronger properties of a BA lead to weaker agenda conditions for achieving the oligarchy result. To provide additional agenda conditions for the oligarchy result, let us introduce the concept of monotonicity (MON) for a BA as follows:
\begin{center} 
(MON) If $\vec{P}(A) \leq \vec{P}'(A)$ and $F(\vec{P})(A)=1$, then  $F(\vec{P}')(A)=1$
\end{center}
where $\leq$ is applied to each component of two vectors.
If we assume MON, we can bypass the need to prove (Fact 1), thereby eliminating the requirement for the agenda to be even-negatable (EN). This is because (Fact 1) is already implied by SYS and MON. The following table illustrates the agenda conditions that are sufficient to achieve the oligarchy result based on different properties of BA:

{\renewcommand{\arraystretch}{1.2}%
\begin{center}
\begin{tabular}{ c||c|c| }
\cline{2-3}
 &  IND &  SYS \\
\hline \hline 
 \multicolumn{1}{ |c||  }{without MON}
& PC, EN     &  NS, EN\\
\hline
 \multicolumn{1}{ |c||  }{with MON}
& PC     &  NS \\
\hline
\end{tabular}
\end{center}

It is noteworthy that the agenda condition required for our oligarchy result is the same as the one for the dictatorship and oligarchy results in judgment aggregation (e.g., \cite{DH10} \cite{DL08}).
In our proof of the if-direction, we extend their counterexamples to our domain in a manner that satisfies UD, ZP, CP, IND, CDC, and CCS: 
the counterexample for a non-path-connected agenda is a minimal extension satisfying MON, as we do not exclude even-negatablility;
the one for a non-even-negatable agenda is an extension satisfying not MON but SYS, as we do not exclude path-connectedness.
So the proof follows a similar structure to those in judgment aggregation, but the ways of extension to construct counterexamples are not trivial---particularly the counterexample for not even-negatable agendas---, and so our proof includes novel ideas that are needed due to the difference between binary and probabilistic beliefs.

\section{The Agenda Condition for the Triviality Result}
\label{sec4}

This section presents and proves our second main result: the agenda condition for the triviality result.
Stronger properties of a BA yield weak agenda conditions. Thus, one might ask whether the agenda condition for the oligarchy result can be weakened, if we add AN.
We will demonstrate that the agendas that yield the triviality result can be characterized by negation-connectedness, which is also the agenda condition for an impossibility result of belief binarization methods as shown in \cite{DL21}.

\begin{definition}[Negation-connected Agenda]
An agenda $\mathcal{A}$ is negation-connected (NC) iff for every contingent issue $A \in \mathcal{A}$ it holds that $A \vDash^{**} \overline{A}$.
\end{definition}
So negation-connectedness means that every issue has a path to its complement. 
According to Proposition 1 in Dietrich \& List (2021), the agenda being negation-connected is equivalent to the agenda being partitioned into subagendas each of which is path-connected, where a subagenda is a non-empty subset of the agenda that is closed under complementation.

%
%

The following lemma will be needed for the proof of the first part of the succeeding theorem.
Part (1) allows us to consider the stronger condition, namely path-connectedness, than negation-connectedness to prove the triviality result. Part (2) will be used when the agenda is path-connected and not even-negatable.

\begin{restatable}[]{lemma}{NCEN}
\label{lem:NC and not EN}
(1) If the triviality result holds---i.e., the only BA on $\mathcal{A}$ satisfying UD, CDC, ZP, CP, IND, and AN is the trivial one---for any path-connected agenda $\mathcal{A}$,
then the same holds for any negation-connected agenda.

(2) If an agenda $\mathcal{A}$ is not even-negatable, then for any minimally inconsistent subset $\mathcal{Y} \subseteq \mathcal{A}$ and any even-sized subset $\mathcal{Z} \subseteq \mathcal{Y}$ it holds that $\mathcal{Y}_{\neg \mathcal{Z}}$ is also minimally inconsistent.
\end{restatable}

The following lemma will be needed for the proof of the second part of the succeeding theorem. This lemma looks technical but it is closely related to the notion of median point in the next section. Indeed, if $\mathcal{H}_0$ is the empty set, then $\bigcap \mathcal{M}$ is the set of all median points where $\mathcal{H}_0$ and $\mathcal{M}$ are defined in the following lemma.

\begin{restatable}[]{lemma}{nnc}
\label{lem:not NC M}
Let $\mathcal{H}_0$ be the set $\{A \in \mathcal{A} \vert \mbox{ } A \vDash^{**} \overline{A} \mbox{ and } \overline{A }\vDash^{**} A \}$.
If $\mathcal{A}$ is not negation-connected, then there is a non empty subset $\mathcal{M} \subseteq \mathcal{A} \setminus \mathcal{H}_0$ such that for any minimally inconsistent set $\mathcal{Y} \subseteq \mathcal{A}$ it holds that $ \vert \mathcal{Y} \cap \mathcal{M} \vert \leq 1$. Furthermore, for any minimally inconsistent set $\mathcal{Y} \subseteq \mathcal{A}$ intersecting $\mathcal{H}_0$ it holds that $ \vert \mathcal{Y} \cap \mathcal{M} \vert= 0$. In addition, for $B \in \mathcal{A} \setminus \mathcal{H}_0$, it holds that $B \in \mathcal{M}$ iff $\overline{B} \notin \mathcal{M}$.
\end{restatable}
Now let us prove the theorem that negation-connectedness is the sufficient and necessary condition for the triviality result.

\begin{restatable}[Agenda Condition for the Triviality Result]{theorem}{agendatri}
\label{thm:Agenda Conditions for Triviality Result}
An agenda $\mathcal{A}$ is negation-connected iff the only BA on $\mathcal{A}$ satisfying UD, ZP, CP, CDC, IND, and AN is the trivial one.

\end{restatable}

The only-if direction of the theorem shows that the triviality result holds if the agenda is negation-connected, which is a generalization of the triviality result. 
The proof suggests further that, if we assume SYS, then non-simplicity (NS) becomes the sufficient condition to obtain the triviality result. In this case, neither EN nor MON is needed, unlike in the case of Theorem \ref{thm:Agenda Oligarchy}, as illustrated in the following table:

\begin{center}
\begin{tabular}{ c||c|c| }
\cline{2-3}
 &  IND &  SYS \\
\hline \hline
 \multicolumn{1}{ |c||  }{with or without MON}
 & NC & NS \\
\hline
\end{tabular}
\end{center}

Compared to the case of the oligarchy result, when we add AN, we obtain the triviality result even under a weaker agenda condition: (i) instead of requiring path-connectedness (PC), negation-connectedness (NC) is sufficient, and (ii) the triviality result holds even when the agenda is not even-negatable (EN). 
The difference mentioned in (i) does not play a role in finding the sufficient condition according to Lemma \ref{lem:NC and not EN}.  However, the necessary condition is not path-connectedness but negation-connectedness. In cases where the agenda is PC and EN, we can apply Theorem \ref{thm:Agenda Oligarchy} since the oligarchy satisfying AN is the trivial one (i.e., the oligarchy with $M=N$). Thus, we only need to focus on the cases where the agenda is PC and not EN. 
When the agenda is assumed to be not EN, we encounter the following difficulty: to show the triviality result, we used (Fact 1), which could be proved if the agenda was assumed to be EN. Our strategy here is to prove a weaker claim than (Fact 1):\\ 
(Fact 1$''$) If $G(\vec{a}) = 1$, then $G(\vec{c}) = 1$ for all $\vec{c} \geq \vert 2\vec{a} -\vec{1} \vert$.\\
The new claim (Fact 1$''$) is weaker than (Fact 1), as it only guarantees that vectors greater than $\vert 2\vec{a} -\vec{1} \vert$ are mapped to 1, rather than all vectors greater than $\vec{a}$.

%

One might ask whether we can apply the proof presented in Dietrich \& List (2021) to our theorem, or vice versa. However, there are differences between the two proofs. On the one hand, we cannot use their proof because, while they deal with probabilistic beliefs, we are dealing with profiles of probabilistic beliefs: 
in particular, for negation-connected agendas in our framework, we can only show (Fact 1$''$) instead of (Fact 1).
On the other hand, since we have not relied on the assumption that $\vert N \vert \geq 2$, our proof can be applied to the context of belief binarization, where $\vert N \vert =1$, and so we can recover their results.


The if-direction gives a counterexample of the triviality result when an agenda is not negation-connected, which implies the agenda being not path-connected. The counterexample presented in Theorem \ref{thm:Agenda Oligarchy} is not applicable in this case because it does not satisfy AN. Moreover, there would be no counterexample if we only assumed an agenda to be not path-connected. This is the reason why we need to weaken path-connectedness to negation-connectedness, even though they fulfill the same role concerning the sufficient agenda condition for the triviality result. 

Our counterexample for a non-negation-connected agenda is an extension of the belief binarization rule proposed in Dietrich \& List (2021).
We extend the rule while maintaining MON, but not minimally, which differs from the way of the extension in Theorem \ref{thm:Agenda Oligarchy}.


\section{The Agenda Condition for the Impossibility Result}
\label{sec5}

Now we will show that the agendas for the impossibility result can be characterized by blocked agendas.

\begin{definition}[Blocked Agenda]
An agenda $\mathcal{A}$ is blocked iff there is an issue $A \in \mathcal{A}$ such that $A \vDash^{**} \overline{A}$ and $\overline{A }\vDash^{**} A$.
\end{definition}

So a blocked agenda contains an issue that has a path to its complement.
Recall that $\mathcal{H}_0$ is defined by the set $\{A \in \mathcal{A} \vert \mbox{ } A \vDash^{**} \overline{A} \mbox{ and } \overline{A }\vDash^{**} A \}$. Then $\mathcal{A}$ is negation-connected iff $\mathcal{H}_0 = \mathcal{A}$, and  $\mathcal{A}$ is blocked iff $\mathcal{H}_0 \neq \emptyset$. If $\mathcal{A}$ is negation-connected, then it is blocked.
%
%
The following definition and lemma will be needed for the succeeding theorem.

\begin{definition}[Median Point]
\label{def:Median Point}
Let $\mathcal{A}$ be an agenda on the set $W$ of possible worlds. A possible world $m \in W$ is a median point iff for any minimally inconsistent subset $\mathcal{Y} \subseteq \mathcal{A}$, it holds that $\vert \{A \in \mathcal{Y} \vert \mbox{ } m \in A \} \vert \leq 1$. 
\end{definition}

So a median point is a possible world that is contained in at most one issue in every minimally inconsistent set.
%
%
It is well-known in judgment aggregation that if a median point is guaranteed to exist, then we can easily construct an anonymous, complete, and consistent judgment aggregator where a median point is thought of as a default collective judgment unless everybody believes the issue being true/false at the median point to be false/true \cite{NP07}. 
The following lemma states that the agenda not being blocked is the necessary and sufficient condition for the existence of a median point.

\begin{restatable}[]{lemma}{median}
An agenda $\mathcal{A}$ is not blocked iff there is a median point.
\end{restatable}
Now let us formulate and prove our last theorem.

\begin{restatable}[Agenda Condition for the  Impossibility Result]{theorem}{agendaim}
\label{thm:Agenda Conditions for Impossibility Result}
 
An agenda $\mathcal{A}$ is blocked iff there is no BA on $\mathcal{A}$ satisfying UD, CP, IND, CCP, and CCS.

\end{restatable}

Indeed, CCS and CCP together are stronger assumptions than CDC. As a result, we obtain the impossibility result more easily, without assuming AN and non-dictatorship, and with a more relaxed agenda condition. The proof demonstrates that by adding SYS, the impossibility result still holds even without CP and even when no agenda condition is assumed---e.g., even when $\mathcal{A}=\{A, \overline{A}\}$. 

The blocked agenda is also the agenda condition for the impossibility results on judgment aggregation with AN in \cite{NP10} and belief binarization in \cite{DL08}. 
Our counterexample for non-blocked agenda is an extension of the counterexample in Dietrich \& List (2018). It is an extension that satisfies MON, but not minimally so. This is the same as the extension in Theorem \ref{thm:Agenda Conditions for Triviality Result}, but different from the one in Theorem \ref{thm:Agenda Oligarchy}. 
Note that the median point $m$ in the proof of this theorem plays the same role as $\mathcal{M}$ in the proof of Theorem \ref{thm:Agenda Conditions for Triviality Result}. The only difference is that $m$ is a possible world and $\mathcal{M}$ is a set of issues. This difference arises from assuming CDC versus assuming CCS and CCP.

%
%
%

%
%
%

\section{Discussion} \label{sec6}

\begin{table}
\begin{center}
\begin{tabular}{ |l|l|} 
 \hline
There is no BA satisfying ... &  Agenda Condition\\
 \hline 
(1) UD, ZP, CP and IND + CDC + Non-oligarchy & path-connected, even-negatable \\
(2) UD, ZP, CP and IND + CDC + AN + Non-triviality & negation-connected \\
(3) UD, CP and IND + CCS and CCP & blocked \\
\hline \hline
There is no judgment aggregator satisfying ...  & Agenda Condition \\
 \hline
(1$'$) UD, ZP, CP and IND + CDC + Non-oligarchy & path-connected, even-negatable \\
(2$'$) UD, ZP, CP and IND + CDC + AN + Non-triviality & negation-connected \\
(3$'$) UD, CP and IND + CCS and CCP + non-dictatorship & path-connected, even-negatable \\
(4$'$) UD, CP and IND + CCS and CCP + AN & blocked \\
 \hline \hline
There is no belief binarization rule satisfying ... & Agenda Condition \\
\hline 
(2$''$) UD, CCS , CP and IND + CDC + Non-triviality & negation-connected \\
(4$''$) UD, CCS, CP and IND + CCP &  blocked \\
 \hline
\end{tabular}
\end{center}
\caption{Classification of Agenda Conditions for Impossibility Results}
\label{tab: Class of agendas generating an impossibility}
\end{table}

All the results in this paper are stated in Table \ref{tab: Class of agendas generating an impossibility}: (1) path-connectedness and even-negatability constitute the exact agenda condition for the oligarchy result; (2) negation-connectedness is for the triviality result; and (3) blockedness is for the impossibility result. These new findings can be compared to the existing characterization theorems in judgment aggregation and belief binarization. 
Regarding (1), it has the same agenda condition as (1$'$) \cite{DL08} and (3$'$) \cite{DH10} in judgment aggregation.
For (2), it is similar to (2$''$) \cite{DL21} in belief binarization, with the difference being the use of ZP instead of CCS for for (2$''$). Since applying our proofs can weaken CCS to ZP, the agenda condition for (2$'$), which has not been discussed in the literature, is also negation-connected because an anonymous and independent judgment aggregator can be viewed as a belief binarization function.
As for (3), it is similar to (4$'$) \cite{NP10} in judgment aggregation and (4$''$) \cite{DL18} in belief binarization.


Let us mention some further research topics. One might think that the rationality norms for collective binary beliefs could be weakened since adhering to deductive closure might be too demanding for group agents. Instead, we could focus on requiring group beliefs to respect consistency or pairwise consistency. By exploring these weaker norms, we can investigate stronger impossibility results.
Furthermore, let us discuss how to obtain possibility results. For this purpose, it is advantageous that binarizing belief aggregation provides a framework that generalizes the problem of judgment aggregation or belief binarization. As in judgment aggregation, we can employ and study premise-based binarizing belief aggregation methods. Alternatively, we can combine an individual belief binarization procedure with judgment aggregation. If we assume that linear or geometric pooling methods are very natural given individual credences, we can apply belief binarization methods to the pooled group credence. Of course, we can also come up with new procedures that cannot be reduced to existing methods.
Ultimately, we should keep in mind that binarizing belief aggregation is an \textit{epistemic} collective decision problem. Therefore, we should be concerned about which methods accurately track the truth. One natural approach would be to investigate belief binarization methods that minimize the expected distance from the truth in light of the group's pooled credence.
In conclusion, binarizing belief aggregation opens a new research area in which various procedures of belief aggregation, different studies on the relation between credences and beliefs, and epistemic decision theory can be combined and explored.

\paragraph*{Acknowledgements}
We would like to express our sincere gratitude to Hannes Leitgeb and Christian List for their invaluable feedback and profound insights. The research conducted by the first author benefited from the generous support of the German Academic Scholarship Foundation and the Alexander von Humboldt Foundation.

\nocite{*}
\bibliographystyle{eptcs}
\bibliography{TARK23W}

\end{document}